\documentclass{svproc}
\usepackage{cite}
\usepackage{url}
\usepackage{multirow}
\usepackage{graphicx}
\usepackage{amsmath, amssymb}
\usepackage{float}

\newcommand{\eq}[1]{(\ref{#1})}
\newcommand{\fig}[1]{Fig.\ref{#1}}

\begin{document}
\mainmatter              
\title{Information Entropy Initialized Concrete Autoencoder for Optimal Sensor Placement and Reconstruction of Geophysical Fields}

\titlerunning{Entropy Initialized Concrete Autoencoder}  
%
\author{Nikita A. Turko\inst{1, \dagger} \and Alexander A. Lobashev\inst{2} \and
Konstantin V. Ushakov\inst{3,1} \and Maxim N. Kaurkin\inst{3} \and Rashit A. Ibrayev\inst{4,3}}
\authorrunning{Nikita A. Turko et al.} 
%
%
\institute{Moscow Institute of Physics and Technology, Dolgoprudny, Russia
\and
Skolkovo Institute of Science and Technology, Skolkovo, Russia
\and
Shirshov Institute of Oceanology, Russian Academy of Sciences, Moscow, Russia
\and
Marchuk Institute of Numerical Mathematics, Russian Academy of Sciences,
Moscow, Russia 
}

\maketitle              

\begin{abstract}

We propose a new approach to the optimal placement of sensors for the problem of reconstructing geophysical fields from sparse measurements. Our method consists of two stages. In the first stage, we estimate the variability of the physical field as a function of spatial coordinates by approximating its information entropy through the Conditional PixelCNN network. To calculate the entropy, a new ordering of a two-dimensional data array (spiral ordering) is proposed, which makes it possible to obtain the entropy of a physical field simultaneously for several spatial scales. In the second stage, the entropy of the physical field is used to initialize the distribution of optimal sensor locations. This distribution is further optimized with the Concrete Autoencoder architecture with the straight-through gradient estimator and adversarial loss to simultaneously minimize the number of sensors and maximize reconstruction accuracy.  Our method scales linearly with data size, unlike commonly used Principal Component Analysis. We demonstrate our method on the two examples: (a) temperature and (b) salinity fields around the Barents Sea and the Svalbard group of islands. For these examples, we compute the reconstruction error of our method and a few baselines. We test our approach against two baselines (1) PCA with QR factorization and (2) climatology. We find out that the obtained optimal sensor locations have clear physical interpretation and correspond to the boundaries between sea currents.



\keywords{concrete autoencoder, optimal sensor placement, information entropy, ocean state reconstruction \\ \\ $\dagger$ \text{Corresponding Author}: \email{turko$@$phystech.edu}}
\end{abstract}
\newpage
\section{Introduction}

The motivation to study optimal sensor placement problems is two-fold: firstly, we want to find locations where to place new sensors or observational stations, secondly, we want to optimize data assimilation methods such as ensemble Kalman filters that use the  singular value decomposition (SVD) and slow-down ocean model simulation.  In real-time ocean forecast systems, a typical number of observations for data assimilation is about 300 observations for temperature and salinity fields and about 5000 observations for altimetry and sea ice concentration \cite{altimetry_assimilation}. With the growing number of observations coming from satellites, it is important to find a balance between the amount of assimilated data and the speed of the data assimilation algorithm. One of the ways to find the optimal amount of data is to solve the problem of estimation of informativity of the measurements.


In this work, we will focus on solving the following problem: suppose we have a data set of historical measurements of some physical field. We want to reconstruct these fields using only a small fraction of the observations while maintaining good reconstruction quality. If we want to select $k$ sensors from $n$ grid nodes, where $k$ lies in the interval $[k_{\min}, k_{\max}]$, then our search space grows as $\sum_{ k = k_{\min}}^{k_{\max}}C^{k}_{n}$, which for large $n$ makes it impossible to use direct combinatorial search by measuring the reconstruction error for each possible placement of sensors. Since the exact solution of the optimal sensor placement problem is associated with laborious combinatorial search, one has to use approximate algorithms.

Classical approximate optimization methods for sensor selection assume that the measurement vector at sparse sensor locations is related to the full observation vector via a linear transformation. The Fisher Information Matrix is used to measure the error between the true and reconstructed fields. The Fisher information matrix can be computed as the inverse covariance matrix of the reconstruction errors. Classical methods can be divided into three main groups depending on the loss function used  \cite{nakai}. The first set of methods maximizes the determinant of the Fisher information matrix or, equivalently, minimizes the determinant of the errors covariance matrix \cite{Saito, Wolf, Kumar}. The second set of methods is based on minimizing the trace of the Fisher information matrix \cite{Krause, Nguyen, Nagata}. In particular, if diagonal error covariance is assumed, the loss function becomes the well-known mean square error (MSE) between the observed and reconstructed fields. The third group of classical methods minimizes the minimum eigenvalue of the Fisher information matrix or, in other words, minimizes the spectral norm of the error covariance matrix \cite{Krause, Nguyen}.

In high-dimensional cases even calculating the error covariance matrix is intractable, for example, if we consider global ocean circulation models with a resolution of $1/4^\circ$ or $1/10^\circ$, the covariance matrix has size up to $10^{6}\times10^{6}$. 
After the introduction of the Gumbel-softmax trick \cite{gumbel_softmax} and the concrete distribution \cite{concrete_dist} which allowed to use traditional gradient methods to learn parameters of discrete probability distributions several deep learning approaches were proposed for optimal sensor placement including concrete autoencodes \cite{concrete_autoencoder, concrete_application}, deep probabilistic subsampling \cite{deep_probabilistic_subsampling} and feature selection network \cite{feature_selection_net}. These deep learning methods are more memory-efficient and don't require costly computation of the covariance matrix.
However, during optimization with help of the Gumbel-softmax trick the long period of exploration of the space of optimal sensor locations is needed. Despite its memory-efficiency, deep learning-based methods may experience slow convergence in case of high-dimensional data. To solve the problem of slow convergence we introduce a prior distribution on the optimal sensor locations by using information entropy.





\section{Information entropy approximation}
\subsection{Information entropy}


\par To apply optimization algorithms consistently in the high-dimensional space of sensor locations, we need to introduce an informative prior distribution. For example, it is advisable to place sensors at points where the physical field has high variability. If we have a non-negative scalar field $V(x,y)$ reflecting the historical variability of the physical field, then the prior distribution becomes
\begin{equation}
    \mathbb{P}(x,y) = \frac{e^{\frac{1}{\tau} V(x,y)}}{\int e^{\frac{1}{\tau} V(x,y)}dx dy}
\end{equation}
where $\tau$ is a temperature parameter that controls the concentration of sensors near the maximum of the variability field $V(x,y)$.

\par The variability of the physical field as a function of spatial coordinates can be estimated as the standard deviation of the values of the physical field, taken along the temporal dimension. This estimate using the historical standard deviation is a special case of information entropy, assuming that at each spatial location we have an independent Gaussian random variable with density
\begin{equation}
     \mathbb{P}(\xi|x,y) = \frac{1}{\sqrt{2 \pi} \sigma(x,y)} e^{-\frac{1}{2}(\frac{\xi - \mu(x,y)}{\sigma(x,y)})^2}
\end{equation}
The entropy of a normally distributed random variable is the logarithm of its standard deviation up to a constant
\begin{equation}
\begin{split}
    H(x,y) = - \int \mathbb{P}(\xi|x,y) \log \mathbb{P}(\xi|x,y) d \xi = \frac{1}{2} + \log(\sqrt{2 \pi} \sigma(x,y)) = \\
    = \log(\sigma(x,y)) + \text{const}.
\end{split}
\end{equation}
Information entropy as a quantitative measure of variability can be generalized to the case of spatially correlated non-Gaussian random variables. One way to implement such an approximation is to use neural networks to approximate the joint probability density of patches taken from a data set of historical values of physical fields.

\begin{figure}[h!]
    \begin{center}
        \includegraphics[width=0.7\textwidth]{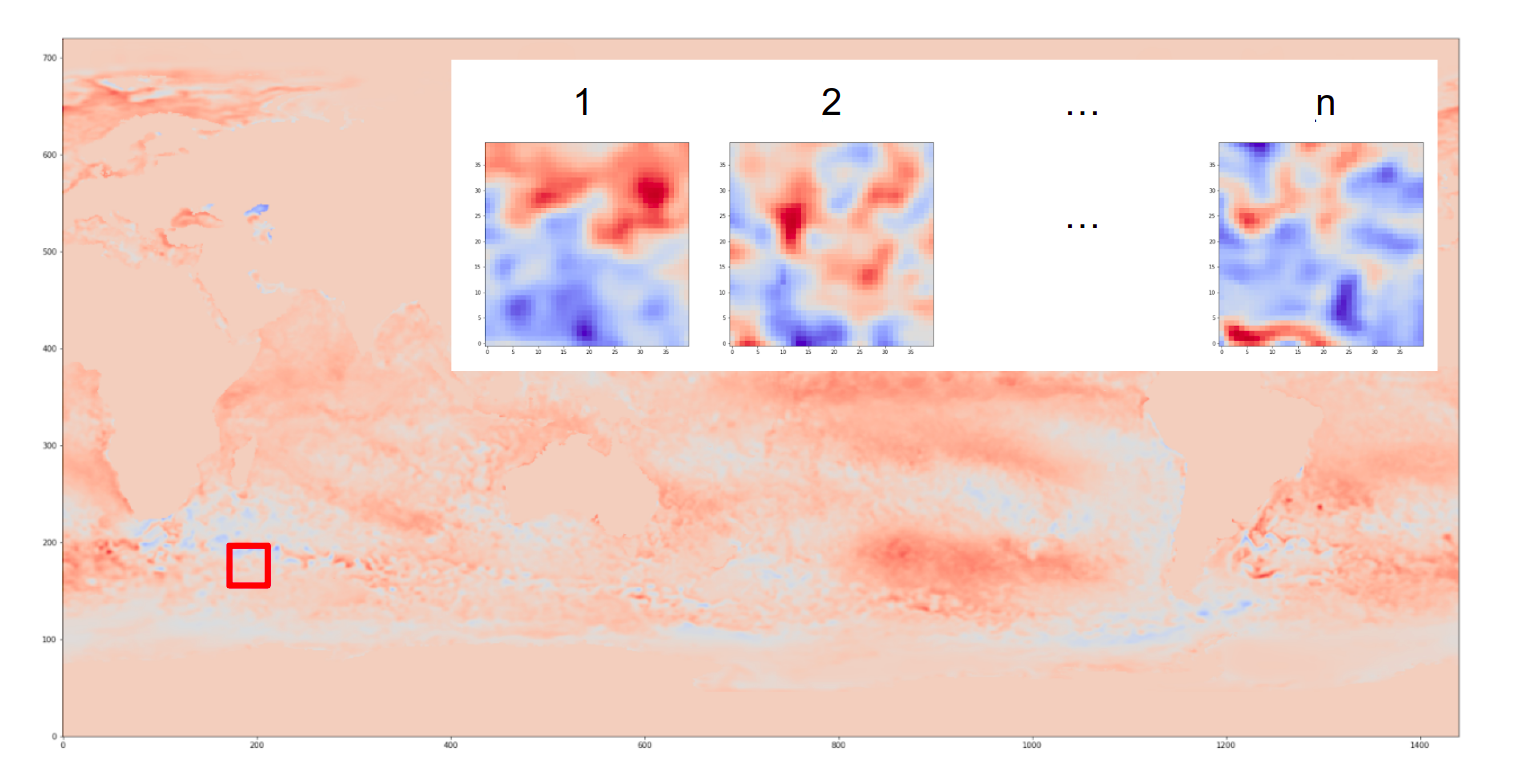}\\
        \caption{Example of patches extracted from sea the surface temperature anomaly field. Indices $1, 2, \dots, n$ correspond to patches taken in the same spatial location at different time moments.}
        \label{patch_examples}
    \end{center}
\end{figure}
Examples of patches are provided in \fig{patch_examples}. We are interested in approximating the conditional density of physical field values in a patch given its center location $P(\xi|x,y)$. The entropy of a physical field could be obtained from the density by the Monte-Carlo approximation
\begin{equation}
\begin{split}
    H(x,y) = - \int \mathbb{P}(\xi|x,y) \log \mathbb{P}(\xi|x,y) d \xi = - \frac{1}{N}\sum_{i=1}^{N} \log \mathbb{P}(\xi_{i}|x,y), \ \xi_{i} \sim \mathbb{P}(\xi_{i}|x,y).
\end{split}
\end{equation}

\subsection{Density estimation using Conditional PixelCNN}

Suppose our set of physical fields is encoded as a set of $L\times L$ images or patches $s_{i}$ cropped from the domain of interest $\mathcal{D}$, each of which is labeled with spatial coordinates of the patch center $\bold{r} = \{r_{1}, \dots, r_{K}\} \in \mathcal{D}$
\begin{equation}
\mathcal{S}^{(\bold{r}_{i}\in \mathcal{D})} = \{s^{\bold{r}_{1}}_{1}, \dots, s^{\bold{r}_{N}}_{N}\}.
\label{ensemble_with_mixed_parameters}
\end{equation}
We want to approximate probability density of physical fields $\mathbb{P}(s^{\bold{r}}|\bold{r})$ conditioned on a vector $\bold{r}$ of spatial coordinates. The image could be represented as the vector of dimension $L\times L$ and the probability density can be expanded as the product of probabilities of individual pixels conditioned on all previous pixels 
\begin{equation}
\mathbb{P}(s^{\bold{r}}|\bold{r}) = \prod_{i=1}^{L\times L} \mathbb{P}([s^{\bold{r}}]_{i}|[s^{\bold{r}}]_{1}, \dots ,[s^{\bold{r}}]_{i-1} , \bold{r}),
\label{pixelcnn_probs}
\end{equation}
where $[s^{\bold{r}}]_{i}$ stands for the i-th pixel of the image $s^{\bold{r}}$ with respect to the chosen ordering.

\begin{figure}[h!]
    \begin{center}
        \includegraphics[width=\textwidth]{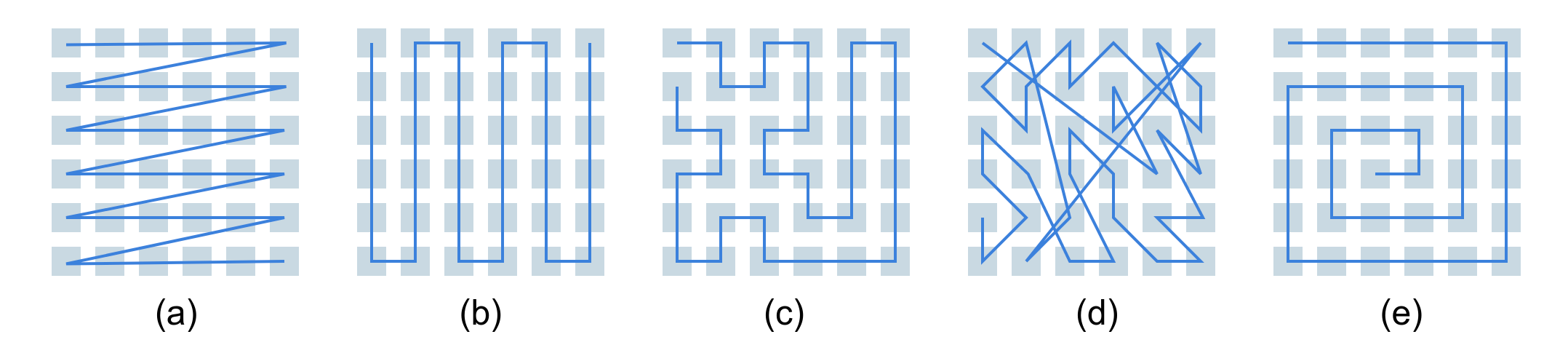}\\
        \caption{(a) Raster ordering; (b) S-curve ordering; (c) Arbitrary continuous ordering; (d) Arbitrary discontinuous ordering. It is favorable to use continuous orderings like (b) and (c) since they don't suffer from the blind spot problem as it does the raster ordering (a) used in the original PixelCNN architecture \cite{oord2016}.  We propose to use spiral ordering (e) since it allows one to compute information entropy at several scales simultaneously.}
        \label{pixel_orderings}
    \end{center}
\end{figure}

\par In Conditional PixelCNN approach \cite{oord2017} all conditional distributions are modeled by a single convolutional network. Auto-regressive structure is achieved by applying masked convolutions. Convolution kernels at every layer are multiplied by a tensor masking all pixels with larger order than the current one. 

In \eq{pixelcnn_probs} we need to specify the particular ordering of pixels in an image. Examples of possible orderings are shown in \fig{pixel_orderings}. We introduce the so-called spiral ordering (e). Spiral ordering allows one to train a Conditional PixelCNN once at the largest patch size $L\times L$ and then obtain information entropy on several spatial scales $L'\times L', \ L' < L$ by taking product only of the first $L'\times L'$ conditional distributions in \eq{pixelcnn_probs}.



\par The network is trained on the dataset $\mathcal{S}^{(\bold{r}_{i}\in \mathcal{D})}$ maximizing probability of observed physical fields or equivalently by minimizing negative log-likelihood
\begin{equation}
L(\theta) = - \sum_{i=1}^{N}\log \mathbb{P}_{\theta}(s_i^{\bold{r}_{i}}|\bold{r}_{i}),
\end{equation}
where $\theta$ is the vector of parameters of Conditional PixelCNN. 

Given a trained network we compute the entropy of a physical field as a function of spatial coordinates by averaging negative log-likelihoods over ensemble of physical fields generated at fixed values of spatial coordinates
\begin{equation}
H(\bold{r}) = - \mathbb{E}_{s} \log \mathbb{P}_{\theta}(s|\bold{r}) = - \frac{1}{\#\{ \bold{r}^{i} :  |\bold{r}^{i}-\bold{r}|<\varepsilon\}} \sum_{|\bold{r}^{i}-\bold{r}|<\varepsilon}\log \mathbb{P}_{\theta}(s_i^{\bold{r}_{i}}|\bold{r}),
\end{equation}
where $\# s$ is the number of elements in a set s.

Computed information entropy field can be used to propose optimal sensor locations by sampling from the distribution
\begin{equation}
    \mathbb{P}(\bold{r}) = \frac{e^{\frac{1}{\tau} H(\bold{r})}}{\int_{\mathcal{D}} e^{\frac{1}{\tau} H(\bold{r})}d\bold{r}}
    \label{sensor_proposal}
\end{equation}
where we set hyperparameter $\tau = 0.2$ for the entropy field measured in nats per computational grid cell.

\section{Optimization of sensors locations and reconstruction of fields}
\subsection{Concrete autoencoder}

The initial sensor locations sampled from the prior distribution \eq{sensor_proposal} need to be further optimized since we also want to find an operator that takes as input sparse measurements and reconstructs the full field from the as small number of measurements as possible. The optimization problem in the space of sensor locations is intrinsically discrete which makes it hard to use effective gradient methods. However, several relaxation-based approaches to solve the problem of differentiation of the samples from a discrete distribution were proposed \cite{concrete_dist, gumbel_softmax}. The Concrete distribution \cite{concrete_dist} was used in the concrete autoencoder \cite{concrete_autoencoder} for differentiable feature selection and sparse sensor placement and has recently been applied in CFD-based sensor placement problems \cite{concrete_application}.

One of the drawbacks of the concrete autoencoder is that we need to apriori specify the desired number of sensors. To automatically define the necessary number of sensors we propose to optimize parameters of the binary mask using a straight-through gradient estimator, i. e. in the backward pass we replace the ill-defined gradients of the step function with the gradients of the identity function.

Our binary mask is parametrized by the scalar field of parameters $w$, and the binary mask representing sensor locations is obtained via a step function
\begin{equation}
    \text{mask}(w) = \text{step}(w)
\end{equation}
\par Straight-through gradient estimator allows us to use a single matrix $w$ of parameters for a different number of sensors. However, the Gumbel-Softmax reparametrization is more efficient at initial exploration the
space of sensor locations. We argue that good initialization of optimal sensor location with
straight-through gradient estimator allows one to achieve similar performance with better-
scaling properties with growing computational grid size and the number of sensors.

Our architecture consists of a trainable binary mask and a reconstructing image-to-image neural network with U-Net architecture with bilinear upsampling. Another version of the reconstructing image-to-image network is a convolutional autoencoder with a discrete binary latent vector.

When training the Concrete Autoencoder $G$ neural network, we find the optimal parameters by minimizing the loss function

\begin{equation}
    \mathcal{L}_{G} =  \mathbb{E}_{S_{\text{full}}}||G(S_{\text {full}} \cdot \text{mask}, w) - S_{\text{full}}||_{L_2} + \lambda \cdot \mathbb{E}|\text{mask}|
\end{equation}
where the function $G$ takes as input the physical field $S_{\text{full}}$ in the entire simulation area, multiplies it component by component by the binary mask $\text{mask}$ and tries to restore the original field. The binary mask must remain as sparse as possible, $\lambda$ is a factor that determines the sparseness of the binary mask. The average value of the average $\mathbb{E}|\text{mask}|$ is proportional to the number of sensors.

\subsection{Concrete autoencoder with adversarial loss}

Once we have a set of pairs from the exact and sparsely measured physical fields (which we obtain using initialization of the binary input mask using information entropy prior), we can solve the inverse problem of recovering the exact physical field from the sparsely measured one by solving the image-to-image translation problem. To do this, we introduce two neural networks following the Pix-2-Pix framework \cite{pix2pix}.

The generator takes a sparsely measured physical field as input and tries to restore the exact one. A network with $\text{U}\text{Net}$ architecture with bilinear upsampling \cite{unet} and with trainable binary input mask is used as a generator $ G: \hat{S} \rightarrow S $.
The discriminator in the space of physical fields tries to distinguish the real exact physical fields from those generated by the generator, $ D: S \times \hat{S} \rightarrow [\mathbb{O},\mathbb{I}] $, where $[\mathbb{O},\mathbb{I}]$ is the set of matrices of size $k \times k$ with entries in the interval $[0,1]$. The specific value of $k$ depends on the size of the input matrix $M$. Discriminator $D$ has PatchGAN architecture used in \cite{pix2pix}.

The loss function of the generator has three main terms
\begin{equation}
    \mathcal{L}_{G} = \lambda_{1} \mathcal{L}_{\text{LSGAN}} + \lambda_{2} \mathcal{L}_{\text{pixel-wise}} + \lambda_{3} \mathcal{L}_{\text{sensors}}
\end{equation}
where we use $\lambda_{1}=10^{-4}$, $\lambda_{2} = 1$ and $\lambda_{3}$ dynamically changes during training from $0$ to $1$. This is done to initially achieve good reconstruction quality and only after to minimize the number of sensors without significant increase of the reconstruction error.

The loss function $\mathcal{L}_{\text{LSGAN}}$ requires that the generator produces physical fields that are indistinguishable by the discriminator from the real fields
$$
\mathcal{L}_{\text{LSGAN}} = \text{MSE}(D(G(\hat{M} )),\mathbb{I}) \equiv ||D(G(\hat{M} )) - \mathbb{I} ||_{L_{2}}
$$

In addition to the requirement for the realism of the generated physical fields, we require an element-by-element correspondence of the restored physical fields with the ground true ones. The $L_{2}$-norm is used to measure pixel-wise error
$$
\mathcal{L}_{\text{pixel-wise}} = ||G(\hat{M})-M)||_{L_{2}}
$$
The control of sensors number is achieved by adding average value of the binary mask to the loss function
$$
\mathcal{L}_{\text{sensors}} = \mathbb{E}||\text{mask}||_{L_{1}}
$$
\par For the discriminator, we require that it distinguish the physical fields produced by the generator from the real ones.
$$
\mathcal{L}_{D} = \frac{1}{2}\left[\text{MSE}(D(G(\hat{M})),\mathbb{O})+\text{MSE }(D(M), \mathbb{I})\right]
$$


\section{Numerical experiments}

The dataset contains daily temperature and salinity fields taken from the INMIO COMPASS global ocean circulation model for 3m and 45m depth near the Svalbard group of islands from 2 Jan 2004 to 31 Dec 2020. The full ocean model dataset was obtained using supercomputer resources of JSCC RAS and INM RAS. Calculation of 17 model years took about 340 hours, using 176 cores for the ocean, 60 cores for ice, and 3 cores for coupler, atmospheric forcing and rivers runoff, so in total  239 cores. The number of processors was chosen so that the time for calculating the integration steps for the ice and ocean models was approximately equal.  This choice leads to more efficient use of computing resources when scaling the model to larger grids.  A detailed study of this issue is presented in \cite{kalnitskii}. The dataset is divided into the train and test sets by taking the first 80\% of the historical values for training and the other 20\% is taken for testing. We train the concrete autoencoder, PCA-QR, and compute daily climate based on the training set. After training, we compute the reconstruction error on the test set and compare all three methods using bias and root mean square error (RMSE) metrics.

All neural networks were trained using the compute nodes with 4 Nvidia Tesla V100 GPUs of the Zhores HPC \cite{zhores}. The information entropy field was computed by averaging across the ensemble of Conditional PixelCNN networks. For every physical field out of four considered, an ensemble of 30 Conditional PixelCNN networks was trained using 10 V100 GPUs. The training of a single PixelCNN takes 1 hour on a single V100 for patch size $16 \times 16$. The training of a single Concrete Autoencoder takes 6 hours on a single V100 for a computational grid $104 \times 284$.

\subsection{Dataset description}
\textbf{INMIO COMPASS Ocean general circulation model.} The system of equations of three-dimensional ocean dynamics and thermodynamics in the Boussinesq and hydrostatic approximations is solved by the finite volume method \cite{bryan} on the type B grid \cite{lebedev_a, lebedev_b, mesinger}. The ocean model INMIO COMPASS \cite{ushakov_ibrayev} and the sea ice model CICE \cite{hunke_lipscomb} operate on the same global tripolar grid with a nominal resolution of $0.25^\circ$. The vertical axis of the ocean model uses z-coordinates on 49 levels with a spacing from 6 m in the upper layer to 250 m at the depth. The barotropic dynamics is described with the help of a two-dimensional system of shallow water equations by the scheme \cite{Killworth}. Horizontal turbulent mixing of heat and salt is parameterized with a background (time-independent) diffusion coefficient equal to the nominal value at the equator and scaled towards the poles proportionally to the square root of the grid cell area.  To ensure numerical stability in the equations of momentum transfer, the biharmonic filter is applied with a background coefficient scaled proportionally to the cell area to the power 3/2 and with the local addition by Smagorinsky scheme in formulation \cite{griffies_hallberg} for maintaining sharp fronts. Vertical mixing is parameterized by the Munk–Anderson scheme \cite{munk_anderson} with convective adjustment performed in case of unstable vertical density profile. On the ocean-atmosphere interface, the nonlinear kinematic free surface condition is imposed with heat, water and momentum fluxes calculated by the CORE bulk formulae \cite{griffies}. Except for vertical turbulent mixing, all the processes were described using time-explicit numerical methods which allow simple and effective parallel scaling. The time steps of the main cycle for solving model equations are equal for the ocean and the ice. The ocean model, within the restrictions of its resolution, implements the eddy–permitting mode by not using the laplacian viscosity in the momentum equations. 

\textbf{The sea ice model CICE v. 5.1.} The simulation regime includes processing of five thickness categories of ice and one of snow, the upwind transport scheme, and the description of melting ponds. The ice dynamics is parameterized with the elastic – viscous – plastic rheology model which requires the subcycle with small time steps for explicit resolving of elastic waves. In the calculation of ice thermodynamics, the zero layer approximation is applied.

\textbf{ERA5 atmospheric forcing.} The ERA-5 reanalysis \cite{hersbach} for the period 2004-2020 was used as the external forcing needed to determine the water and momentum fluxes on the ocean-atmosphere and ice-atmosphere interfaces. Wind speed at 10 m above sea level, temperature and dew point temperature at 2 m were transmitted to the ice-ocean system every 3 hours. In addition, the accumulated fluxes of incident solar and long-wave radiation, precipitation (snow and rain) are read with the same period.

\subsection{Baselines}

\subsubsection{Climate.}

The simplest baseline in ocean modeling, reconstruction, and forecasting is climate interpolated in time to the correct
date. We calculated our climate values on the train set for each day in a year, according to the formula 
\begin{equation}
    S^{climate}(i,j,d) = \frac{1}{N^{years}} \sum_{y = 1}^{N^{years}} S(i,j,y,d),
\end{equation}
where $S(i,j,y,d)$ - the value of physical field with coordinates $(i, j)$ at day number $d = \{1, 2, ..., 365 \}$ in year $y$ from train set, $N^{years}$ - number of years with day $d$ in train set.

\subsubsection{PCA-QR.}  Principal component analysis (in the geophysics literature also known as Proper Orthogonal Decomposition or the method of Empirical Orthogonal Functions) with pivoted QR decomposition is a common baseline for sparse sensor placement \cite{manohar, Wolf, Kumar}. This method approximately minimizes the determinant of the covariance matrix of reconstruction errors to find optimal sensor locations. The main assumption of PCA is that the joint density of the data and the low-dimensional latent variables has the form
\begin{equation}
    \mathbb{P}(x, z| \theta) = \mathbb{P}(x|z, \theta)\mathbb{P}(z) = \mathcal{N}(x|\mu+W z, \sigma^{2}I) \cdot \mathcal{N}(z|0, I),
\end{equation}
where $\theta = \{ \mu, W, \sigma\}$ represents tunable parameters. These parameters could be found from maximization of the marginal likelihood of the observed dataset
\begin{equation}
    \mathbb{P} (x| \theta)= \int \mathbb{P}(x, z| \theta) dz \rightarrow \underset{\theta}{\max} 
\end{equation}
which in the limit of $\sigma \rightarrow 0$ is equivalent to computation of the eigenvalues of the covariance matrix of the data. By applying the QR decomposition with column pivoting to the matrix of eigenvalues $W^{T}$ we will obtain matrices $P, Q, R$ such that 
\begin{equation}
    W^{T}P = QR,
\end{equation}
where $Q$ is an orthogonal matrix, $R$ is an upper triangular matrix and $P$ is a permutation matrix. The permutation matrix $P$ contains optimal sensor positions and plays the role of an operator which extracts a measurement vector $M$ at optimal sensor locations $M = P X$. Then the full field could be reconstructed from the sparse measurements $M$ as
\begin{equation}
    X^{rec} = W M (W^{T} P)^{-1}
\end{equation}
\subsection{Evaluation metrics}

Evaluation metrics used in this study were based on GODAE OceanView Class 4 forecast verification framework \cite{class4}. 
$Bias$ demonstrates correspondence between mean forecast and mean observation. For the analysis the  $Bias$ was calculated in every spatial location $(i,j)$ with averaging along the time using eq. (\ref{fig:bias_field}) and for each time moment in the test set with averaging across the spatial coordinates using eq. (\ref{fig:bias_history})

\begin{equation}
    Bias(i,j) = \frac{1}{\#\{ \tau \in Test Set\}} \sum_{\tau \in Test Set} (S^{recon}(i,j,\tau)-S^{ref}(i,j,\tau)),
    \label{fig:bias_field}
\end{equation}
where $S^{recon}(i,j,\tau)$ - reconstructed values of a physical field at a point with coordinates $(i, j)$ and at time moment $\tau$, $S^{ref}(i,j,\tau)$ - original values of a physical field in the same point.

\begin{equation}
    Bias(\tau) = \frac{1}{N^i}\frac{1}{N^j} \sum_{i = 1}^{N^i}  \sum_{j = 1}^{N^j}(S^{recon}(i,j,\tau)-S^{ref}(i,j,\tau)),
    \label{fig:bias_history}
\end{equation}
where $N^i \cdot N^j$ - total number of computational cells for a physical field.

The second metric used is the Root Mean Square Error ($RMSE$). It was calculated in every grid point $(i,j)$ with averaging along the time dimension  using eq. (\ref{fig:rmse_field}) and for each time moment in test set with averaging along the spatial dimensions using eq. (\ref{fig:rmse_history})
\begin{equation}
    RMSE(i,j) = \sqrt{\frac{1}{\#\{ \tau \in Test Set\}} \sum_{\tau \in Test Set} (S^{recon}(i,j,\tau)-S^{ref}(i,j,\tau))^2}
    \label{fig:rmse_field}
\end{equation}

\begin{equation}
    RMSE(\tau) = \sqrt{\frac{1}{N^i}\frac{1}{N^j} \sum_{i = 1}^{N^i}  \sum_{j = 1}^{N^j}(S^{recon}(i,j,\tau)-S^{ref}(i,j,\tau))^2}
    \label{fig:rmse_history}
\end{equation}

The scalar metric presented in the Tab. \ref{table_with_results} was calculated by taking the median along the time dimension in the $RMSE(\tau)$ time series.

\subsection{Results}

\begin{figure}[h!]
    \centering
    \includegraphics[width=0.9\textwidth]{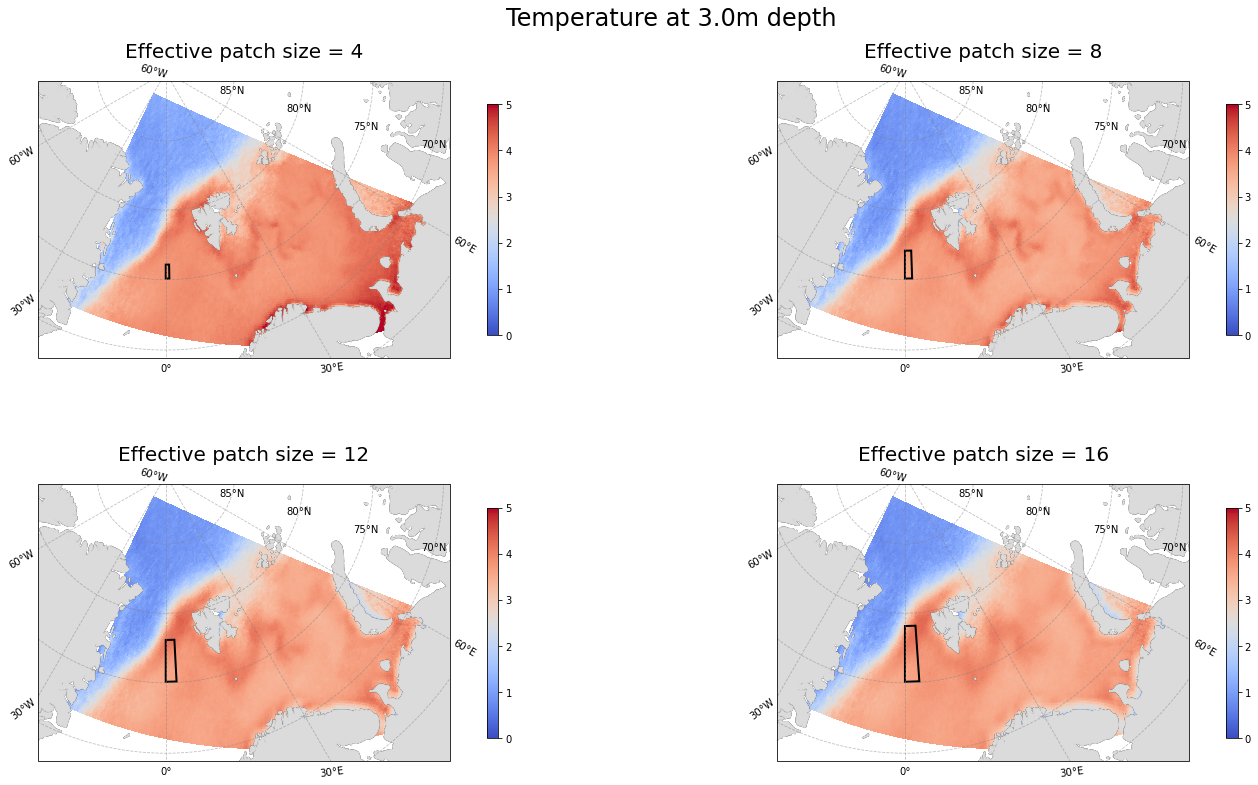}
    \caption{Smoothed ensemble average information entropy of temperature field at 3 meter depth calculated by application of PixellCNN to test data set. The black outline shows the size of the smoothing region with different side size: (a) $1\times1^\circ$ (b) $2\times2^\circ$ (c) $3\times3^\circ$ (d) $4\times4^\circ$}
    \label{effective_patches}
\end{figure}

\textbf{Information entropy.} \fig{effective_patches} shows the information entropy of the temperature field on the under-surface sea layer, calculated from the test dataset. The color indicates the value of information entropy in nats per computational grid cell. The value characterizes the average level of "uncertainty" inherent to the variable's possible outcomes. A small value means that we had more a priori information about the temperature at this point than in a point with higher information entropy.  The black contour highlights the areas of various sizes from $1\times1^\circ$ to $4\times4^\circ$ along which smoothing is performed. Smoothing allows one to remove anomaly and noise, but the smoothing scale should be chosen carefully because the land boundaries and hydro-physical features of the region should be preserved. We chose the size $2\times2^\circ$ , because it allows one to meet the specified requirements.

\begin{figure}[h!]
    \centering
    \includegraphics[width=0.9\textwidth]{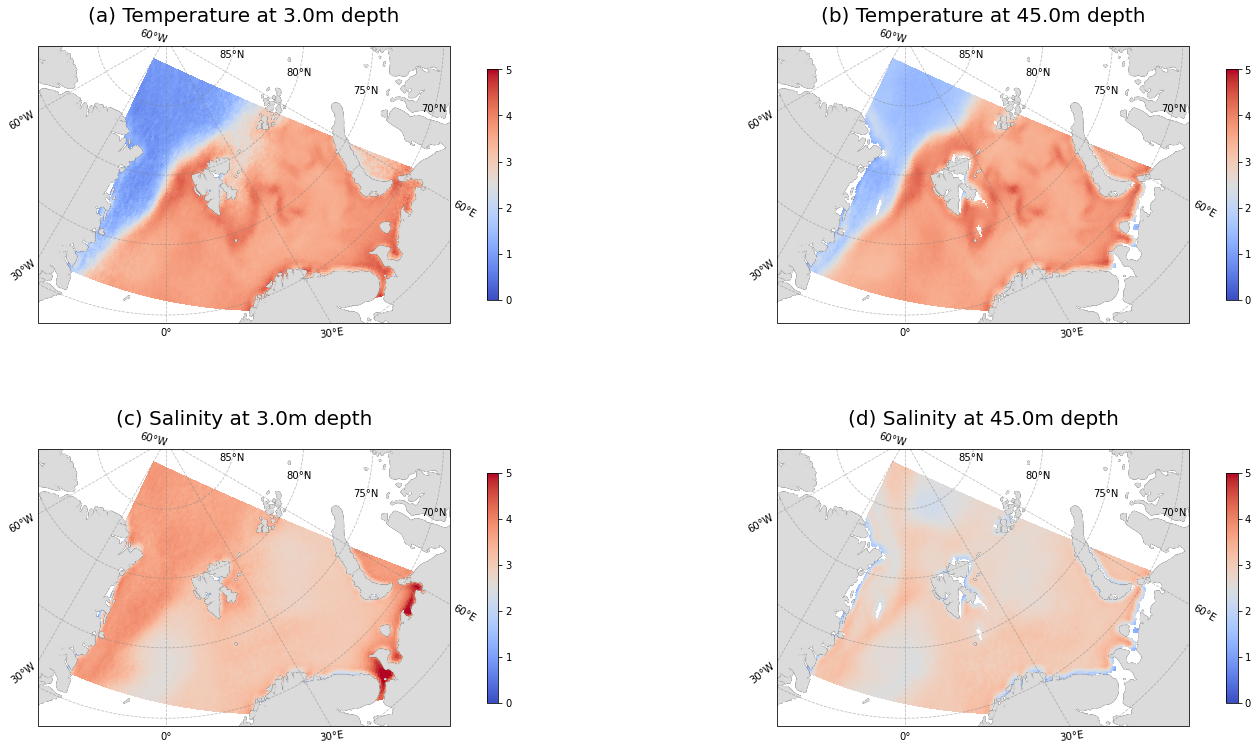}
    \caption{Smoothed ensemble mean information entropy of geophysical fields: temperature at (a) 3 meter and (b) 45 m depth; salinity at (c) 3 meter and (d) 45 m depth.}
    \label{information_entropy}
\end{figure}

\fig{information_entropy} shows the information entropy fields calculated from the test set for temperature and salinity fields at under-surface and depth layers. The result obtained is consistent with the hydrophysics of the region under consideration. Temperature fields show low uncertainty northwest of Svalbard. Because most of the time the surface of the ocean is covered with ice with constant close to the ice formation temperature. And south of Svalbard and in the Barents Sea, high information entropy values are observed. These are areas of mixing of relatively warm North Atlantic waters and cold waters from the Arctic Ocean. For salinity, the situation is significantly different in the northwestern part of the region, since although the area of ice cover varies to a lesser extent, the growth and melting occur constantly, which in turn changes the salinity each time, and as a result, there is a high degree of uncertainty. High values of information entropy on the 3m depth along the border with land and estuaries in the southeastern part of the region are observed. This also corresponds to physical considerations: river freshwater reduces salinity, and in shallow water, the movement of water masses has a high degree of variability due to the influence of wind.

\begin{table}[h!]
\caption{Test set reconstruction errors} 
\centering
\begin{tabular}{ |p{4.5cm}||p{3cm}|p{2cm}|p{2cm}|  }

 \hline
 \multicolumn{4}{|c|}{Temperature 3m} \\
 \hline
 Method & Number of sensors & MED(Bias) & MED(RMSE) \\
 \hline
 Climate                &   0       &   -0.19     &   0.98\\
 PCA with QR            &   77      &    0.13     &   1.03\\
 Concrete Autoencoder   &   77      &   -0.07     &   \textbf{0.73}\\
 \hline
 \hline
 \multicolumn{4}{|c|}{Temperature 45m} \\
 \hline
 Method & Number of sensors & MED(Bias) & MED(RMSE) \\
 \hline
 Climate                &   0       &   -0.09     &   0.88\\
 PCA with QR            &   72      &    0.11     &   1.10\\
 Concrete Autoencoder   &   72      &   -0.05     &   0.83\\
 Concrete Autoencoder LSGAN  &   42      &   0.07     &   \textbf{0.73}\\
 \hline
  \hline
 \multicolumn{4}{|c|}{Salinity 3m} \\
 \hline
 Method & Number of sensors & MED(Bias) & MED(RMSE) \\
 \hline
 Climate                &   0       &    0.58     &   0.84\\
 PCA with QR            &   57      &   -0.03     &   0.66\\
 Concrete Autoencoder   &   57      &    0.05     &   \textbf{0.53}\\
 \hline
  \hline
 \multicolumn{4}{|c|}{Salinity 45m} \\
 \hline
 Method & Number of sensors & MED(Bias) & MED(RMSE) \\
 \hline
 Climate                &   0       &    0.59     &   0.72\\
 PCA with QR            &   61      &    0.02     &   \textbf{0.30}\\
 Concrete Autoencoder   &   61      &    0.26     &   0.41\\
 \hline
\end{tabular}
\label{table_with_results}
\end{table}

\begin{figure}[h!]
    \centering
    \includegraphics[width=0.9\textwidth]{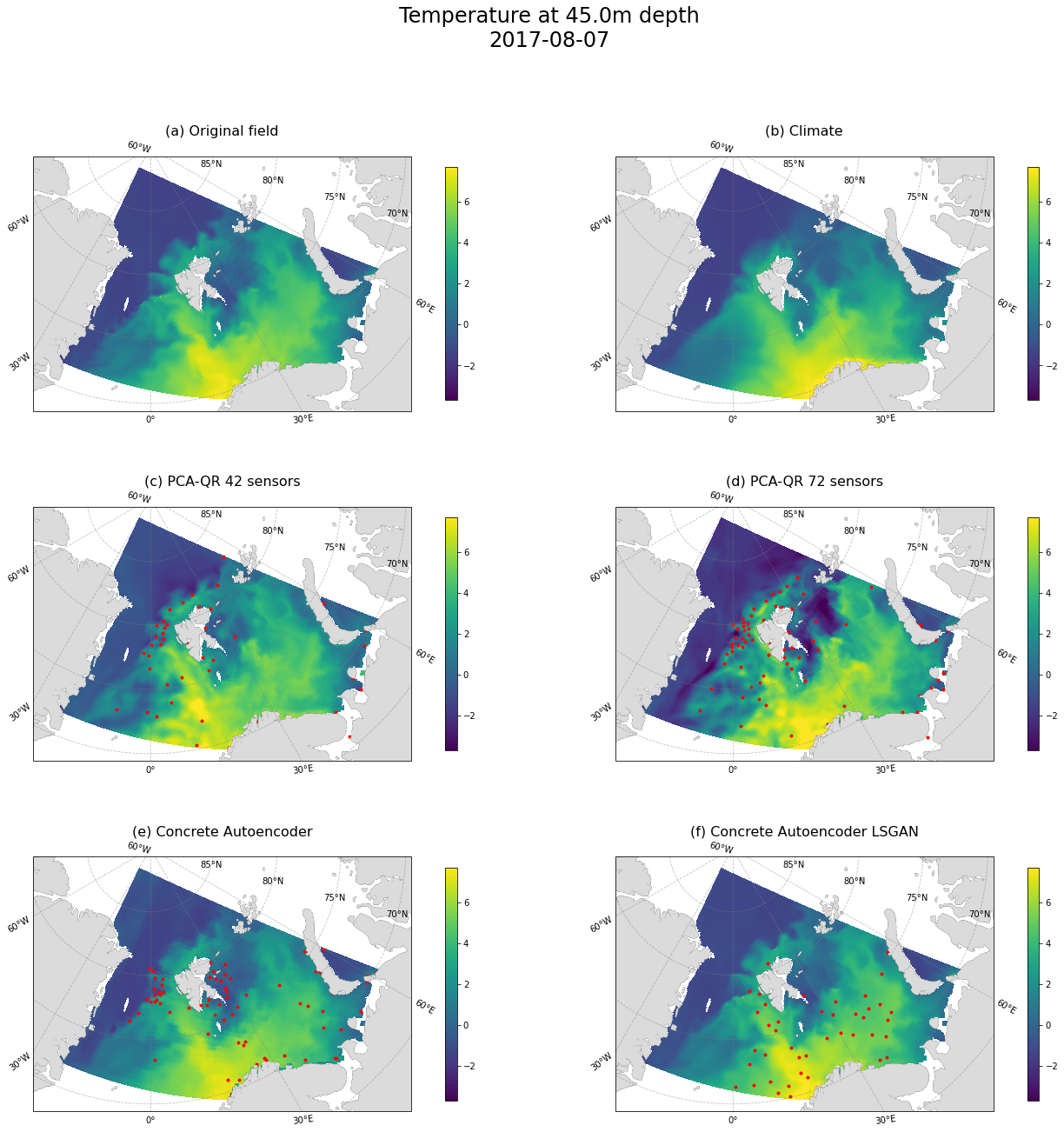}
    \caption{Examples of reconstructed fields. (a) Original temperature field from modeled data at 45 m depth at 00:00 $7^{th}$ August, 2017. The temperature field reconstructed from measurements of the original field at the points marked in red using methods: (b) Climate at 00:00 $7^{th}$ August, no sensors. (c) PCA-QR, 42 measurements. (d) PCA-QR, 72 measurements. (e) Concrete Autoencoder, 72 measurements. (f) Concrete Autoencoder with LSGAN, 42 measurements.}
    \label{reconstructed_fields}
\end{figure}

\begin{figure}[h!]
    \centering
    \includegraphics[width=0.75\textwidth]{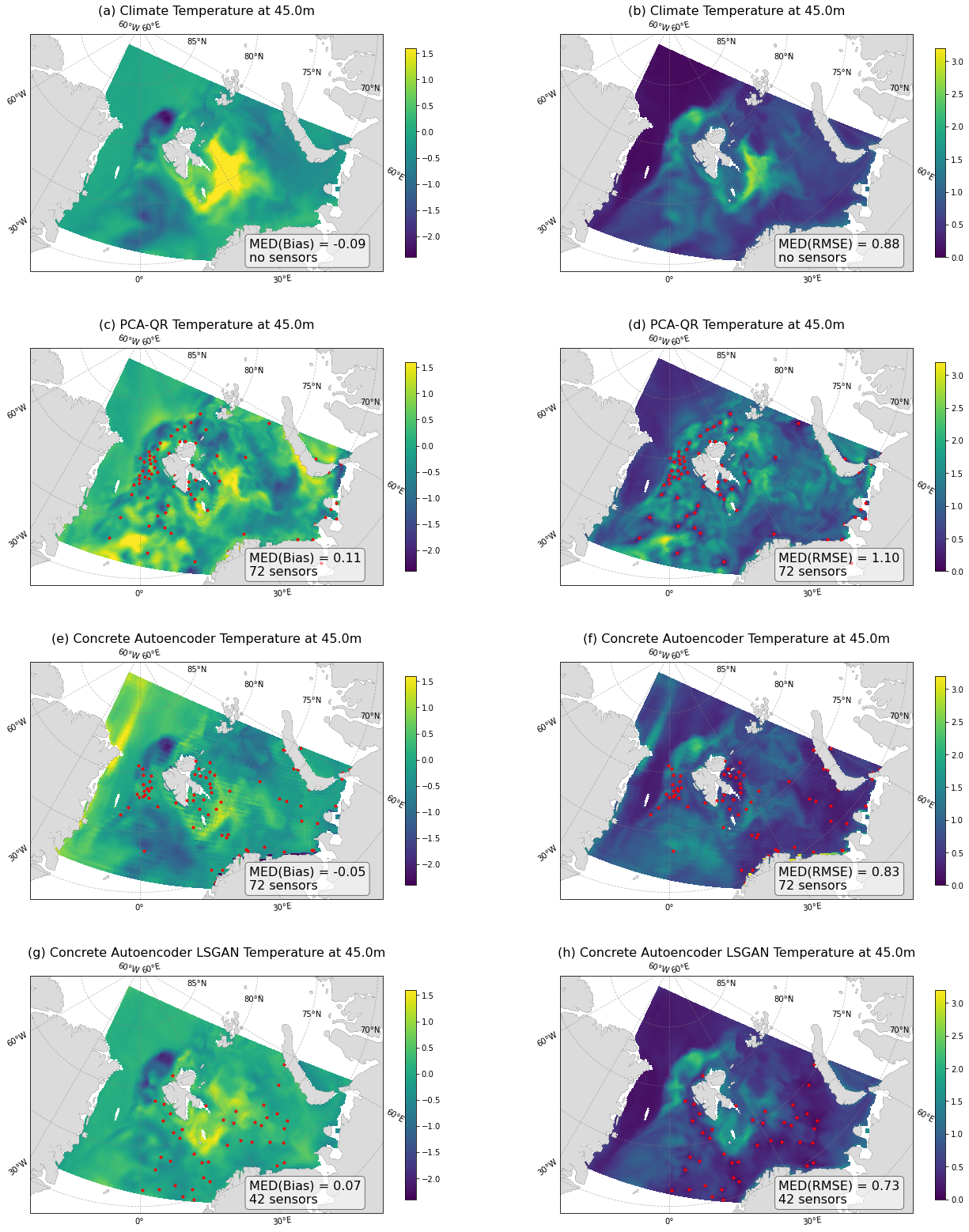}
    \caption{Temperature field reconstruction based on measurements accuracy against original model data. Bias/RMSE temperature reconstruction at depth 45m by methods: (a)/(b) Climate, (c)/(d) PCA-QR, (e)/(f) Concrete Autoencoder, (g)/(h) Concrete Autoencoder with LSGAN. At the left corner of each plot shown the time median field value and number of measurements.}
    \label{bias_rmse_fields}
\end{figure}

\begin{figure}[h!]
    \centering
    \includegraphics[width=0.75\textwidth]{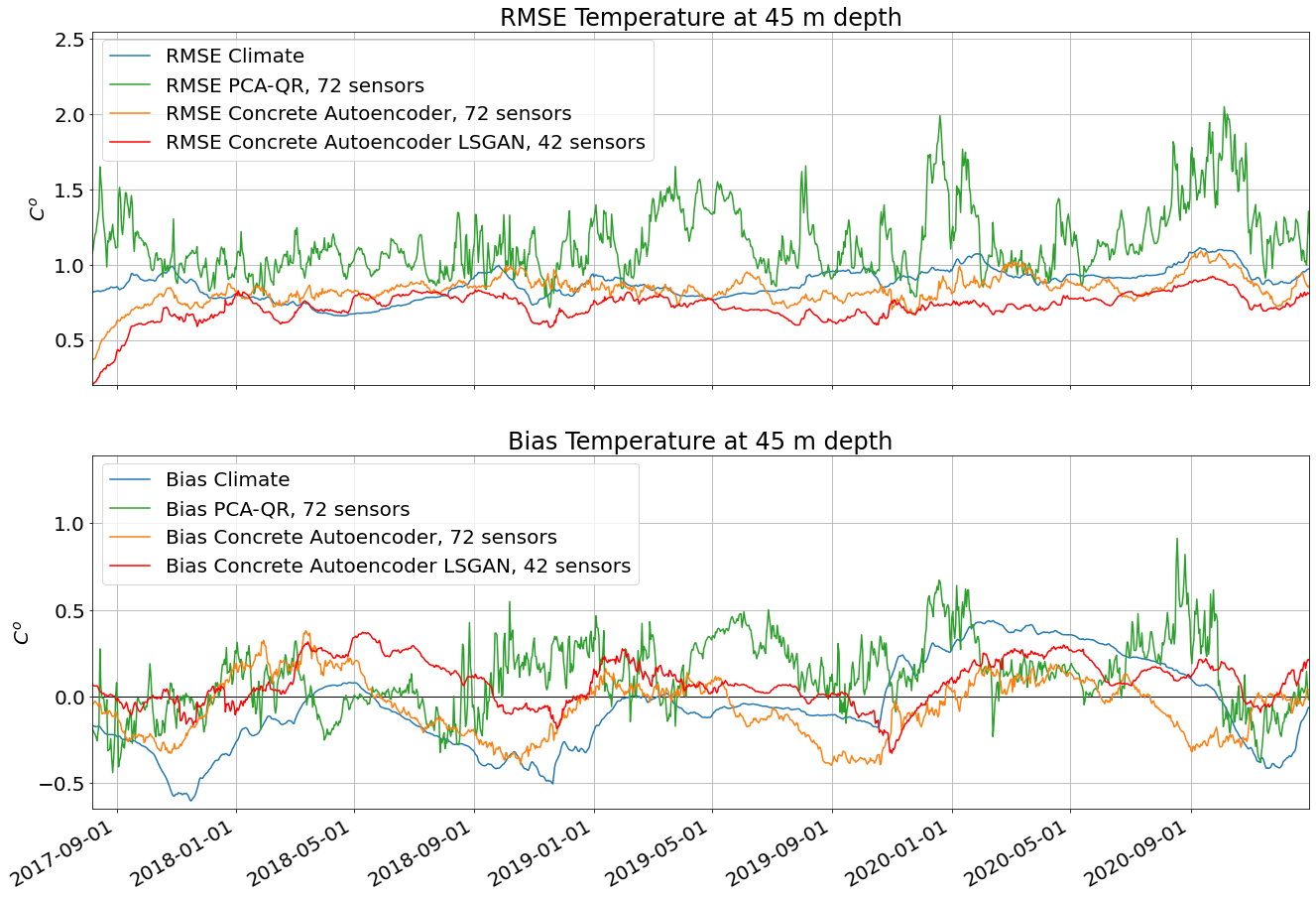}
    \caption{Temperature field reconstruction accuracy against original model data. RMSE/Bias time series based on data from the test dataset.}
    \label{bias_rmse_history}
\end{figure}

\textbf{Reconstruction Accuracy.} Examples of reconstructed fields are shown in \fig{reconstructed_fields}. The accuracy metrics calculated by the formulas (\ref{fig:bias_field}) and (\ref{fig:rmse_field}) are shown in the \fig{bias_rmse_fields}. The median $MED$ value was calculated from the time series of $Bias$ and $RMSE$ using the formulas (\ref{fig:bias_history}) and (\ref{fig:rmse_history}). Corresponding historical time series is shown in the \fig{bias_rmse_history}.

The dipole error structure for climate \fig{bias_rmse_fields} errors fields (a) and (b) shows the presence of the interannual variability in the dataset, this is also confirmed by the absence of such a structure for calculating the error on the training set. There is significant noise in field reconstruction errors using PCA-QR, which is caused by the use of high-order modes that learn to resolve mesoscale eddies on the training set and, due to the presence of interannual variability, make a significant contribution to the increase in reconstruction errors. In the Concrete Autoencoder (e) error picture, artifacts are visible along the parallels, which can be removed by replacing MSE with LSGAN loss function. It should be noted that there is an excessive number of sensors along the coast for the PCA-QR and Concrete Autoencoder methods. This can be explained by the fact that, through atmospheric forcing, the ocean feels the influence of the earth, whose temperature has a higher annual variability due to lower heat capacity. Thus, Concrete Autoencoder with LSGAN better reconstructs the instantaneous physical field using fewer sensors.

Finally, the \fig{bias_rmse_history} shows the graphs of $Bias$ and $RMSE$ versus time, calculated using the formulas \eq{fig:bias_history} and \eq{fig:rmse_history} on the test sample. From the shape of the $Bias$ climate error curve, one can quantify the scale of interannual variability. It can also be seen from the graphs that the PCA-QR method shows the worst result in terms of field reconstruction accuracy due to over-fitting to the training set.  Concrete Autoencoder with LSGAN loss demonstrates the best restoration accuracy, and the $Bias$ graph shows a clear correlation with the climate error up to the spring of 2019. In general, the influence of seasonal variability is expressed in a cyclical change in $Bias$ for all methods, which can be partially eliminated if the physical field anomaly relative to the climate is initially used for training and testing.


\section{Conclusion}
We proposed a method for optimal sensor placement and reconstruction of geophysical fields. The method consists of two stages. At the first stage, we estimate the variability of the physical field as a function of spatial coordinates by approximating the information entropy of patches of field values through the Conditional PixelCNN. 
We proposed a modification to the Conditional PixelCNN architecture by introducing spiral ordering. This new ordering allows one to compute entropy at several scales using the network trained with a relatively large patch size. In our experiments, we observe that the patch size of 8 corresponding to the square $2\times 2$ degrees produces the information entropy field that captures spatial characteristics of the sea currents near the Svalbard group of islands in the most consistent way.

In the second stage, the information entropy field is used to initialize the binary mask of the Concrete Autoencoder, and corresponding sensor locations are further optimized to maximize the reconstruction accuracy of the physical fields and minimize the total number of sensors. The reconstruction accuracy was measured by two losses. Firstly, we try to mean square error as a loss. Secondly, we use adversarial LSGAN loss. 

As a result, we observe that the proposed method outperforms baselines and that the addition of LSGAN loss improves reconstruction accuracy. PCA-QR overfits on the training data more severely than the Concrete Autoencoder both with MSE and LSGAN loss. We observe that LSGAN loss gives the lowest reconstruction RMSE and Bias out of the considered methods while maintaining a spatially well-distributed set of sensors and avoiding clustering of sensors in some particular areas of the considered region.



Such an approach could reconstruct the synoptic variability from a limited number of measurements, and at the same time allows one to explore the inter-annual variability.

\section*{Acknowledgments}

This research was funded by the state assignment of IO RAS, theme FMWE-2021-0003 (analysis of the temperature and salinity fields near the Svalbard group of islands, final experiments and assessment of reconstruction accuracy for the considered optimal sensor placement methods), and by the BASIS Foundation, Grant No. 19-1-1-48-1 (development of the information entropy approximation scheme and initial experiments with the concrete autoencoder by A. L.). The authors acknowledge the use of Zhores HPC \cite{zhores} for obtaining the results presented in this paper. The ocean model dataset was obtained using supercomputer resources of JSCC RAS and INM RAS.


%
%

\bibliography{main}
\bibliographystyle{unsrtnat}

\end{document}